\def\year{2022}\relax
\documentclass[letterpaper]{article} 
\usepackage{aaai22}  
\usepackage{times}  
\usepackage{helvet}  
\usepackage{courier}  
\usepackage[hyphens]{url}  
\usepackage{graphicx} 
\usepackage{caption} 
\usepackage{algorithm}
\usepackage[normalem]{ulem}
\usepackage{newfloat}
 \usepackage{float}
\usepackage{listings}
\usepackage{caption}

\usepackage{amsmath}

\title{Evaluating Biomedical BERT Models for Vocabulary Alignment at Scale in the UMLS Metathesaurus}
\author{
    Goonmeet Bajaj\textsuperscript{\rm 1}\textsuperscript{\rm *}, Vinh Nguyen\textsuperscript{\rm 2}, Thilini Wijesiriwardene\textsuperscript{\rm 3}, Hong Yung Yip\textsuperscript{\rm 3}, Vishesh Javangula\textsuperscript{\rm 4}, Srinivasan Parthasarathy\textsuperscript{\rm 1}, Amit Sheth\textsuperscript{\rm 3}, Olivier Bodenreider \textsuperscript{\rm 2}\textsuperscript{\rm *}
}
\affiliations {
    \textsuperscript{\rm 1} The Ohio State University\\
    \textsuperscript{\rm 2} National Library of Medicine \\
    \textsuperscript{\rm 3} AI Institute,  University of South Carolina \\
    \textsuperscript{\rm 4} George Washington University \\
    \textsuperscript{\rm *} Corresponding authors \\
    \{bajaj.32, parthasarathy.2\}@osu.edu,
    vinh.nguyen@nih.gov, obodenreider@mail.nih.gov, \{thilini, amit\}@sc.edu, hyip@email.sc.edu, visheshj123@gwu.edu
}

\usepackage{bibentry}
\usepackage[dvipsnames]{xcolor}
\usepackage{hyperref}

\begin{document}

\maketitle

\begin{abstract}
The current UMLS (Unified Medical Language System) Metathesaurus construction process for integrating over 200 biomedical source vocabularies is expensive and error-prone as it relies on the lexical algorithms and human editors for deciding if the two biomedical terms are synonymous. Recent work has aimed to improve the Metathesaurus construction process using a deep learning approach with a Siamese Network initialized with BioWordVec embeddings for predicting synonymy among biomedical terms.  Recent advances in Natural Language Processing such as Transformer models like BERT and its biomedical variants such as BioBERT small, BioBERT large, BlueBERT, and SapBERT with contextualized word embeddings have achieved state-of-the-art (SOTA) performance on downstream tasks. These techniques are therefore logical candidates for a synonymy prediction task as well.

In this paper, we evaluate different approaches of employing biomedical BERT-based models in two model architectures: (1) Siamese Network, and (2) Transformer for predicting synonymy in the UMLS Metathesaurus. We aim to validate if these approaches using the BERT models can actually outperform the existing approaches. In the existing Siamese Networks with LSTM and BioWordVec embeddings, we replace the BioWordVec embeddings with the biomedical BERT embeddings extracted from each BERT model using different ways of extraction. In the Transformer architecture, we evaluate the use of the different biomedical BERT models that have been pre-trained using different datasets and tasks. Given the SOTA performance of these BERT models for other downstream tasks, our experiments yield surprisingly interesting results: (1) in both model architectures, the approaches employing these biomedical BERT-based models do not outperform the existing approaches using Siamese Network with BioWordVec embeddings for the UMLS synonymy prediction task, (2) the original BioBERT large model that has not been pre-trained with the UMLS outperforms the SapBERT models that have been pre-trained with the UMLS, and (3) using the Siamese Networks yields better performance for synonymy prediction when compared to using the biomedical BERT models.
\end{abstract}

\section{Introduction}

The Unified Medical Language System (UMLS) \cite{bodenreider2004unified} is a biomedical terminology integration system that includes over 200 source vocabularies\footnote{\url{https://www.nlm.nih.gov/pubs/techbull/mj21/mj21_umls_2021aa_release.htmll}}, including CPT, ICD-10, MeSH, and SNOMED CT.
The UMLS Metathesaurus construction process organizes synonymous terms from these source vocabularies into \textit{concepts}. The Ontology Alignment Evaluation Initiative\footnote{\url{http://oaei.ontologymatching.org/}} (OAEI) uses three well-defined vocabularies (NCI, FMA, and SNOMED CT) from the Metathesaurus for their ontology alignment task \cite{euzenat2011ontology}. Unlike the ontologies used by OAEI, not all vocabularies in the UMLS are well-defined or represented as ontologies. Therefore, when we refer to Metathesaurus construction process we use the phrase \textit{vocabulary alignment} instead of ontology alignment.

The Metathesaurus construction process uses a lexical similarity model and semantic pre-processing to determine synonymy. Human editors determine the final set of synonymous terms. The large scale and diversity of the Metathesaurus make the construction process very challenging, tedious, and error-prone for human editors. To assist the UMLS Metathesaurus construction process, Nguyen et al. introduce the UMLS Vocabulary Alignment (UVA) task, or synonymy prediction \cite{nguyen2021biomedical}. The synonymy prediction task takes in terms (or ``atoms") as input to determine synonymy among them. The authors design and train a Siamese network to predict if two atoms from UMLS are synonymous \cite{nguyen2021biomedical, nguyenadding}. The Siamese network is initialized using BioWordVec embeddings that are learned using fastText \cite{bojanowski2017enriching} to encode atom strings. The authors use the Manhattan distance to compute the (dis)similarity in the final output representations from the Siamese network. Synonymous pairs can be predicted using different similarity thresholds. We describe this approach in more detail in Section \ref{relatedwork}.

Given the success of Transformer models in Natural Language Processing (NLP) \cite{devlin2018bert, vaswani2017attention}, we evaluate different approaches of employing biomedical BERT-based models in two model architectures: (1) a Siamese Network \cite{nguyen2021biomedical, nguyenadding}, and (2) a Transformer model \cite{devlin2018bert} for predicting synonymy in the UMLS Metathesaurus.
We first evaluate different feature extraction techniques to replace BioWordVec embeddings with BERT embeddings in the current state-of-the-art Siamese Networks used for synonymy prediction \cite{nguyen2021biomedical, nguyenadding}. Second, we evaluate the use of the Transformer architecture using the biomedical BERT models for synonymy prediction. In particular, we use nine different biomedical BERT models:
BioBERT \cite{lee2020biobert},
BioBERT Large \cite{lee2020biobert},
BlueBERT \cite{peng2019transfer},
SapBERT \cite{liu2020self},
UMLSBERT \cite{michalopoulos2020umlsbert},
BioBERT + SapBERT \cite{liu2020self},
BlueBERT + SapBERT \cite{liu2020self},
UMLSBERT + SapBERT \cite{liu2020self},
VanillaBERT + SapBERT \cite{liu2020self}.
We aim to validate if these approaches using the BERT models can actually outperform the existing approaches for synonymy prediction.

\textbf{Contributions.} Given the SOTA performance of these BERT models for other downstream tasks, our experiments yield surprisingly interesting results:
(1) in both model architectures, the biomedical BERT-based models do not outperform the existing approaches using a Siamese Network with BioWordVec embeddings for the UMLS synonymy prediction task,
(2) the original BioBERT large model that has not been pre-trained with the UMLS outperforms the SapBERT models that have been pre-trained with the UMLS, and (3) using the Siamese Networks yields better performance for synonymy prediction when compared to using the biomedical BERT models.


\section{Background: Knowledge Representation in the UMLS Metathesaurus}

\begin{table}
\resizebox{\columnwidth}{!}{
\begin{tabular}{|l|l|l|l|l|}
\hline
Tuple & String & Source & AUI & CUI \\ \hline
$t_1$ & Headache & MSH & A0066000 & C0018681 \\ \hline
$t_2$ & Headaches & MSH & A0066008 & C0018681 \\ \hline
$t_3$ & Cranial Pains & MSH & A1641924  & C0018681   \\ \hline
$t_4$ & Cephalodynia & MSH & A26628141 & C0018681  \\ \hline
$t_5$ & Cephalodynia & SNOMEDCT\_US & A2957278 & C0018681  \\ \hline
$t_6$ & Headache (finding) & SNOMEDCT\_US & A3487586 & C0018681 \\ \hline
\end{tabular}
}
\caption{Examples of synonymous atoms from a Metathesaurus concept with associated identifiers}
\label{example_triples}
\end{table}

The UMLS Metathesaurus links terms and codes between health records, pharmacy documents, and insurance documents \cite{bodenreider2004unified}.
The Metathesaurus consists of several building blocks, including atoms and concepts (clusters of synonymous atoms).
Each atom is a term from a specific source vocabulary and each concept is a cluster (or grouping) of atoms. All atoms in the UMLS Metathesaurus are assigned a unique identifier (AUI). 
When the same term appears in different source vocabularies, the individual terms are assigned separate AUIs. Table \ref{example_triples} contains examples of synonymous atoms and the various types of identifiers assigned to each respective atom for a particular concept (i.e. C0018681). For example, the term “Cephalodynia” appearing in both MSH and SNOMEDCT\_US has different AUIS as shown in Table \ref{example_triples}: “A26628141” and “A2957278” respectively.
Additionally, the strings “Headache” and “Headaches” have different AUIs because of the lexical variation (see Table \ref{example_triples}). Finally, each concept (cluster of synonymous terms) in the Metathesaurus is labelled with a unique identifier (CUI).
To recap, every atom (or term) has an unique identifier (AUI) is linked to a single string $STR$ and belongs to a single concept $CUI$.

\section{Problem Formulation}

To formulate the UMLS Metathesaurus construction process as a machine / deep learning task, Nguyen et al. \cite{nguyen2021biomedical, nguyenadding} view synonymy prediction as a similarity task. The task is to identify synonymous atoms by measuring the (dis)similarity among pairs of atoms. Finding synonymous atoms pairs is comparable to finding a cluster of synonymous atoms. Of note, the authors refrain from treating this as a classification task, because it is unfeasible to train a classifier to predict one out of 4.28 million classes (i.e. concepts) present in the 2020AA UMLS release.

A machine-learning model should be able to identify the (a)synonymy among atoms are that lexically:

\begin{itemize}
\item similar but are not synonymous, e.g., “Lung disease and disorder” versus “Head disease and disorder”
\item dissimilar but are synonymous, e.g., “Addison’s disease” versus “Primary adrenal deficiency”
\end{itemize}

We maintain the same problem definition proposed by \cite{nguyen2021biomedical}. The synonymy prediction task is defined as follows. Let $T$ $=$ $(S_{STR}, S_{SRC}, S_{AUI})$ be the set of all input tuples in the Metathesaurus where:

$S_{STR}$ is the set of all atom names,

$S_{SRC}$ is the set of all source vocabulary names, and

$S_{AUI}$ is the set of all atom unique identifiers.

\noindent
Consider four sample tuples from Table 1:

$t_1$ = (“Headache”, ``MSH", ``A0066000")

$t_3$ = (“Cranial Pains”, “MSH”, ``A1641924")

$t_4$ = (“Cephalodynia”, “MSH”, ``A26628141")

$t_5$ = (“Cephalodynia”, “SNOMEDCT\_US”, ``A3487586").

The tuples shown here consist of ($str$, $src$, $aui$), where $str$ is the original string of the term from the source vocabulary ($src$), and $aui$ is the unique atom identifier.

Let ($t_i$, $t_j$) be a pair of input tuples, where $i \neq j$ and each tuple is initialized from a different source vocabulary in the form of ($str$, $src$, $aui$). Let
$f: T \times T \rightarrow {0, 1}$ be a prediction function that maps a pair of input tuples to either 0 or 1. If $f(t_i, t_j) = 1$, then the atom strings ($str_i$, $str_j$) from $t_i$ and $t_j$ are synonymous.

\section{Dataset}

 We thank Nguyen et al. for sharing the training data used in their work \cite{nguyen2021biomedical, nguyenadding}. The dataset is created using the 2020AA release of the UMLS Metathesaurus and only contains English terms from active source vocabularies. There are approximately 27.9M synonymous pairs (positive samples) in the UMLS and the approximately $10^{14}$ pairs of non-synonymous atom (negative samples).
 The ratio of negative samples to positive samples is large because most atoms do not share a CUI. To create a better class balance between the negative and positive samples, the authors reduce the negative samples to approximately 170M \cite{nguyen2021biomedical}. Additionally, Nguyen et al. \cite{nguyen2021biomedical, nguyenadding} create different dataset splits based on the lexical similarity and the number of negative and positive training pairs. In this work, we use the \textit{ALL} split of the dataset used in \cite{nguyen2021biomedical, nguyenadding}. The \textit{ALL} dataset contains the following splits:
 \begin{itemize}
     \item TOPN\_SIM: negative pairs with the highest similarity
     \item RAN\_SIM: random negative pairs having some similarity
    \item  RAN\_NOSIM: random negative pairs having no similarity
 \end{itemize}
 The training and testing datasets are mutually exclusive and do not contain overlapping examples. For our study, we use the $ALL$ dataset and Table \ref{dataset_stats} contains the dataset statistics. We refer the readers to Section 4.2 of \cite{nguyen2021biomedical} for a complete description of the dataset generation process.

\begin{table}[]
\resizebox{\columnwidth}{!}{
\begin{tabular}{@{}llll@{}}
\hline
\textbf{Type} & \textbf{Negative Examples} & \textbf{Positive Examples} & \textbf{Total} \\ \hline
Training and Validation & 170,075,628       & 22,324,834        & 192,400,462 \\
Testing                 & 167,454,653       & 5,581,209         & 173,035,862 \\  \hline
\end{tabular}
}
\caption{Dataset Statistics}
\label{dataset_stats}
\end{table}

\section{Related Work}
\label{related_work}

\subsection{Siamese Networks for UVA}

\begin{figure}[t]
\includegraphics[width=8cm]{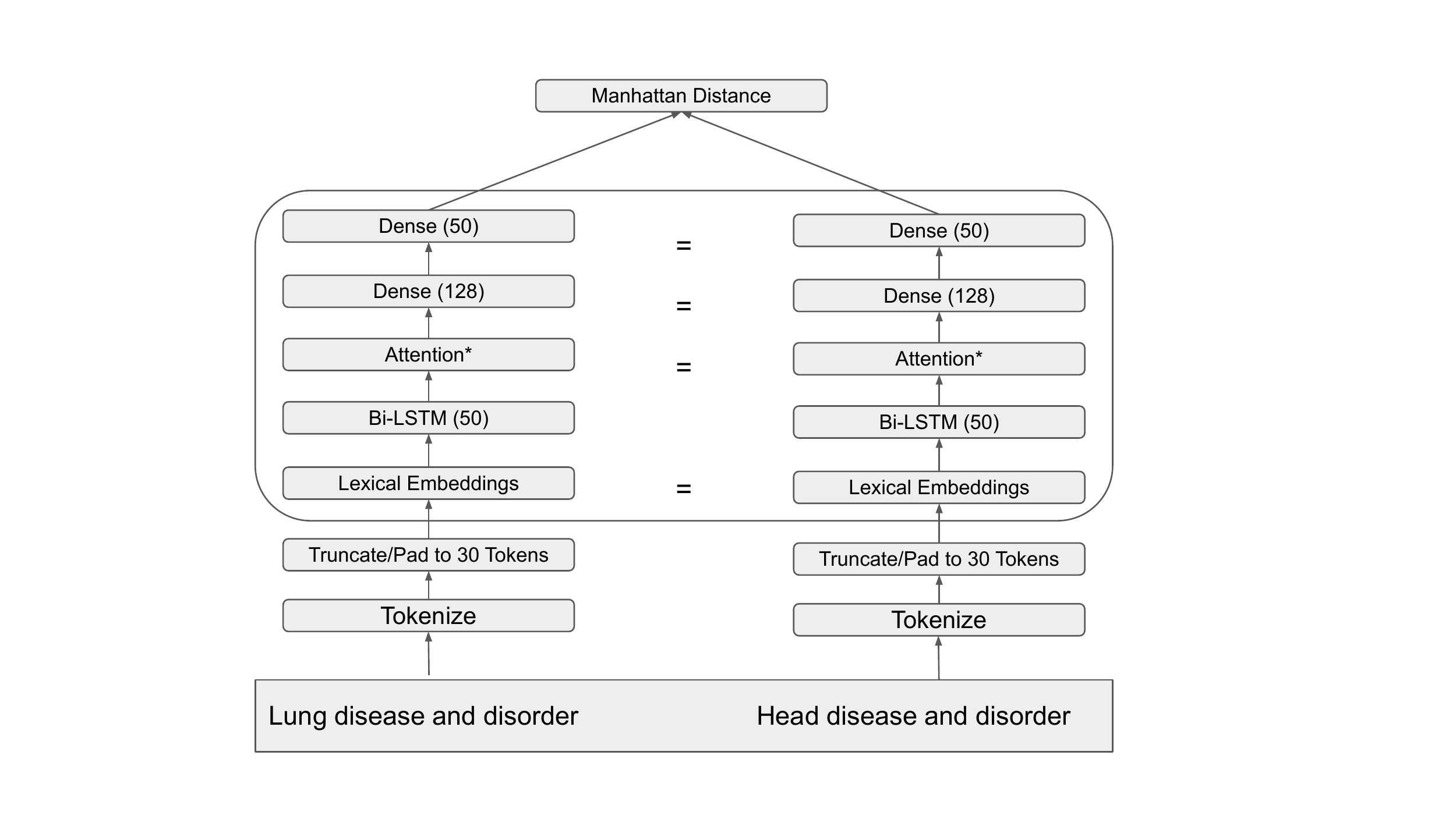}
\caption{Siamese Model used for Synonymy Prediction. Nguyen et al. use BioWordVec embeddings \cite{nguyen2021biomedical, nguyenadding}, whereas, we replace the BioWordVec embeddings with embeddings extracted from domain specific BERT Models.}
\label{siamese_network_fig}
\end{figure}

Nguyen et al. \cite{nguyen2021biomedical} assess the similarity of atoms using lexical features of the atom strings ($str$). The authors design a Siamese Netowrk that inputs a pair of atom strings, ($str_i, str_j$), where  $i \neq j$, and outputs a similarity score between 0 and 1, $sim{(str_i, str_j)} \in [0, 1]$.
The inputs are pre-processed, then sent through an embedding layer initialized with BioWordVec embeddings \cite{zhang2019biowordvec}. The word embeddings are then fed into Bidirectional Long Short Term Memory (Bi-LSTM) layers to learn the semantic and syntactic features of the atoms through as a sequence of tokens. The outputs from the Bi-LSTMs are then fed into two consecutive dense layers consisting of 128 hidden units and 50 respectively. The learned representation for each atom are then fed into a Manhattan distance similarity function, $exp( - ||LSTM_A - LSTM_B ||_1) \in [0, 1]$ to determine the similarity. In their follow-up work \cite{nguyenadding}, Nguyen et al. add an attention layer that improves the precision of the network for synonymy prediction by +3.63\% and decreases recall by -1.42\%. Figure~\ref{siamese_network_fig} displays the Siamese Network architecture. The asterisk next to the attention layer indicates that the additional layer is only used in Nguyen et al.'s follow up work \cite{nguyenadding}.
Given the success of Transformer models for different NLP tasks, the objective of this work is to investigate the performance of context-aware embeddings extracted using different methods from various domain-specific BERT models with the Siamese networks introduced in \cite{nguyenadding, nguyen2021biomedical}. In this work, we replace the BioWordVec embeddings with embeddings extracted from biomedical BERT models using different feature extraction techniques. In Section \ref{approach}, we outline our experimental setup.

\subsection{Contextualized Word Representations and Biomedical BERT Models}

Distributed representations (e.g. word2vec, fastText) provide a single embedding for lexically similar but semantically dissimilar words \cite{mikolov2013distributed, pennington2014glove, bojanowski2017enriching}.
Recent advances in Natural Language Processing (NLP) have led to better performing contexualized embeddings that are learned using Transformer models (e.g. BERT) \cite{vaswani2017attention, devlin2018bert}. Transformer models produce contextualized word representations that are informed by the surrounding words in the input. Additionally, Transformers handle long range dependencies in sequences entirely through self-attention instead of sequence-dependent RNNs \cite{lin2017structured}. Embeddings extracted from Transformer models have outperformed word2vec based embeddings on several NLP tasks \cite{devlin2018bert, vaswani2017attention}. 


Bidirectional Encoder Representations from Transformers (BERT) is a key technical innovation that applies the attention mechanism found in Transformers to language models. Unlike previous efforts in language modelling, which looked at text sequences in a unidirectional manner, BERT processes text sequences bidirectionally and learns a deeper sense for the context and flow of a language \cite{devlin2018bert}. The BERT architecture is designed to provide such contextualized representations. In order to achieve the contextual embeddings, BERT models are trained using two self-supervised training tasks: Masked Language Model (MLM) and Next Sentence Prediction (NSP) (refer BERT paper). Current research shows these models can be pre-trained on large domain-specific corpora and fine-tuned on smaller domain-specific tasks to achieve better performance on downstream tasks \cite{lewis2020pretrained}. Biomedical NLP research follows this trend and has shown that BERT-based models such as BioBERT \cite{lee2020biobert}, BlueBERT \cite{peng2019transfer}, SapBERT \cite{liu2020self} trained on domain-specific datasets outperform models that use more traditional word embeddings generated from models like word2vec and fastText. In Section \ref{sec:biomedical_BERT_variants}, we provide more details about the different Transformer models used in our work.

\section{Biomedical BERT Variants}
\label{sec:biomedical_BERT_variants}

In this section, we explain the differences between the domain-specific BERT variants used in this study. Table \ref{bert_model_comparisons} compares the different biomedical BERT models used in this paper.

\begin{table}[]
\resizebox{\columnwidth}{!}{
\begin{tabular}{@{}llllll@{}}
\hline
\textbf{Model\_Type} & \textbf{Embed. Dim.} & \textbf{Vocab Size} & \textbf{Token Size} & \textbf{\# of Param} & \textbf{\# of Param. w. Attention} \\ \hline
BioWordVec & 200 & 268,158,600 & - & 268,221,858  &  268,221,778\\
BioBERT (+ SapBERT) & 768 & 28,996 & 13,230,336 & 13,407,194 & 13,407,114  \\
BioBERT\_Large (Cased) & 1024 & 58,996 & 28,530,688 & 28,758,666  & 28,758,746  \\
BlueBERT & 1024 & 30,522 & 25,358,336 & 25,586,314 & 25,586,394 \\
SapBERT & 768 & 30,522 & 21,035,520 & 21,212,298 & 21,212,378  \\
UMLSBERT (+ SapBERT) & 768 & 28,996 & 13,230,336 & 13,407,114 & 13,407,194  \\
BlueBERT+ SapBERT & 768 & 30,522 & 19,018,752 & 19,195,530 & 19,195,610 \\
VanillaBERT + SapBERT & 768 & 30,522 & 19,018,752 & 19,195,530 & 19,195,610 \\ \hline

\end{tabular}
}
\caption{Comparison of Siamese Networks initialized with embedding from different biomedical BERT models. The columns indicate the embedding dimension, vocabulary size, number of tokens, number of parameters of Siamese Network, and the number of parameters of the Siamese Network with Attention Layer.}
\label{bert_model_comparisons}
\end{table}

\noindent
\textbf{BioBERT:}
BioBERT is initialized from BERT pre-trained on Wikipedia (2.5 billion words) and Books Corpus (0.8 billion words) \cite{lee2020biobert}. This BERT model is then pre-trained on biomedical domain data consisting of PubMed Abstracts (4.5 billion words) and PMC Full-text articles (13.5 billion words). Then the pre-trained model was used in several Biomedical NLP tasks such as Biomeidcal Named Entity Recognition (BioNER), BioRE and question answering \cite{lee2020biobert}.
For this study, we use BioBERT-Base v1.1, which has 768 hidden units for the embedding layer, and BioBERT-Large v1.1 (trained with a custom vocabulary), which has 1024 hidden units for the embedding layer.


\noindent
\textbf{BlueBERT:} BlueBERT is initialized with BERT weights provided by \cite{devlin2018bert} and further pre-trained with biomedical corpus (PubMed abstract with ~4,000M words) and clinical notes corpus (MIMIC-III with ~500M words). Two versions of BlueBERT are released consistent with BERT-Base and BERT-Large
models trained with 5M steps on the PubMed corpus and 0.2M steps on the MIMIC-III corpus. In our work, we use BlueBERT-Large trained on both PubMed and MIMIC-III datasets.
\newline
\noindent
\textbf{SapBERT:} SapBERT provides the current SOTA results for 6 medical entity linking (MEL) bench-marking datasets. SapBERT is trained on the UMLS with 4M+ concepts and 10M+ synonyms from over 150  vocabularies including MeSH, SNOMED CT, RxNorm. SapBERT is trained using a SOTA metric
learning objective inspired by visual recognition, for learning from the positive and negative pairs of the UMLS.

\noindent
\textbf{BioBERT + SapBERT, BlueBERT + SapBERT, BlueBERT + SapBERT, UMLSBERT + SapBERT, VanillaBERT + SapBERT:} The SapBERT authors pre-train additional variants of SapBERT that are initialized using different BERT variants.
\newline
\noindent
\textbf{UMLSBERT:} UmlsBERT is initialized with the pre-trained
Bio\_ClinicalBERT model \cite{alsentzer2019publicly} and further pre-trained with the MLM task on the MIMIC-III dataset. The authors modify the pre-training task in two ways: 1) by introducing an additional semantic type embedding, 2) modify the MLM task by replacing the 1-hot vector that corresponds to the masked word, with a binary vector to indicate which words share them same CUI as the masked word.

\begin{figure}[t]
\includegraphics[width=8cm]{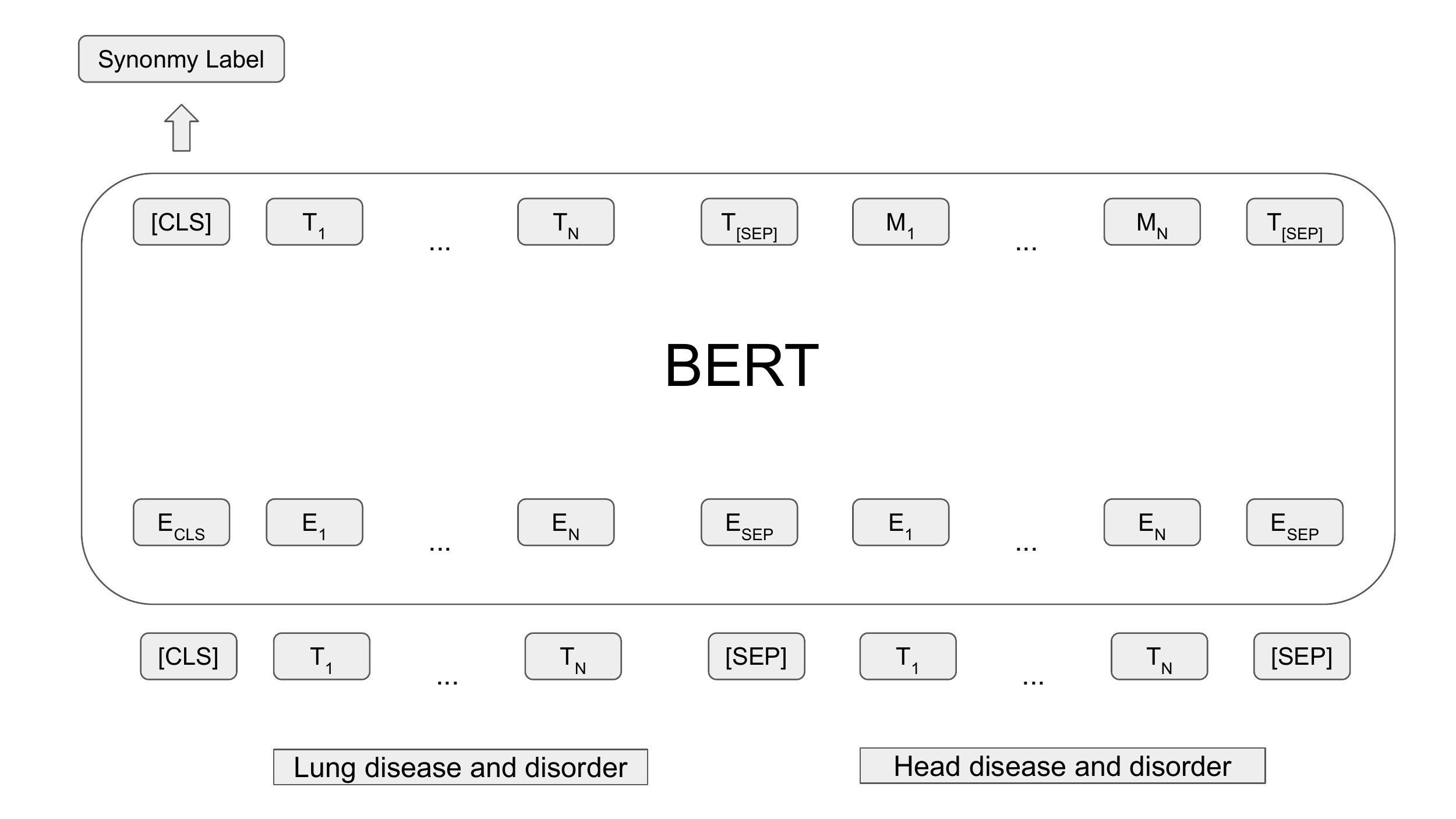}
\caption{BERT Model for Synonymy Prediction}
\label{bert_model_synonymy}
\end{figure}

\section{Approach}
\label{approach}
In this section, we first explain our experimental setup to investigate the performance of the Siamese networks using embeddings extracted from BioBERT, BioBERT Large, BlueBERT, SapBERT, UMLSBERT, BioBERT + SapBERT, BlueBERT + SapBERT, UMLSBERT + SapBERT, and VanillaBERT + SapBERT.
Next, we explain our setup to investigate the performance of the different domain-specific BERT models for the UVA task using the BERT architecture.

\subsection{UVA with Siamese Network}
BioWordVec embeddings are generated using the fastText model \cite{bojanowski2017enriching} on the PubMed text corpus and MeSH. As mentioned, embeddings extracted from Transformer models (e.g. BERT) used to train different deep learning models have achieved state-of-the-art (SOTA) performance on different down-stream tasks because these embeddings are context-dependent. For example, syntactic features are captured better using the middle layers and semantic features at the latter layers of the model \cite{jawahar-etal-2019-bert}. Additionally, the original inventors of BERT share that extracting embeddings from different layers leads to variations in performance on the down-stream task \cite{devlin2018bert}.

We use this insight from \cite{jawahar-etal-2019-bert, liu-etal-2019-linguistic}, to experiment weather the Siamese Network model introduced in \cite{nguyen2021biomedical} can benefit from contextually aware embeddings extracted from biomedical BERT models. We extract different token embeddings from different biomedical BERT models. Different layers of BERT carry different information \cite{jawahar-etal-2019-bert, liu-etal-2019-linguistic, devlin2018bert}. Therefore, we investigate which token embeddings and which layers lead to better performance when used to initialize the Siamese Network.
We extract two sets of embeddings from each model: 1) embeddings from the last layer, and 2) embeddings from the average of the last four layers. Additionally, we use the three different types of token embeddings: 1) the first occurrence of the token in the dataset, 2) the last token in the dataset, 3) the average embedding of each occurrence of the token in the dataset. It is important that we investigate which token embedding is appropriate because the BERT models generate different token embeddings for each token based on the context from the input atom string. Of note, we only use the atom string to extract token embeddings because all vocabularies have this characteristic in common. We tokenize the atom strings using the BERT tokenizer and the respective vocab from each biomedical BERT model.




\subsection{UVA with Transformer Network}

To understand the usability of the domain-specific BERT Models, we use pre-trained models for the UVA task. These experiments allow us to benchmark the performance of the different domain-specific BERT models for the UVA task. Of note, we do not fine-tune these BERT models because the size of our training data.

For the UVA, task we use the following input format to process the atom strings: A [CLS] token is inserted at the beginning of the first atom string $str_i$, followed by a [SEP] token, followed by the second atom string $str_j$, followed by a final [SEP] token to indicate the end of the sequence. The input is then processed through the BERT model and an output of 0 (synonymous) or 1 (not synonymous) is predicted.

\section{Experimental Setup}

In this section, we provide details for our two sets of experiments: 1) feature extraction for the Siamese Network and 2) UVA with transformer networks. We run all experiments using a High Performance Computing (HPC) cluster. We use the following BERT models for both experiments: BioBERT, BioBERT Large, BlueBERT, SapBERT, UMLSBERT, BioBERT + SapBERT, BlueBERT + SapBERT, UMLSBERT + SapBERT, and VanillaBERT + SapBERT.







\subsection{Feature Extraction for the Siamese Network}

To understand the performance of the different embeddings extracted from the various BERT models, we train the Siamese Network end to end. Our experimental setup is similar to \cite{nguyen2021biomedical, nguyenadding}. We refrain from changing the experimental setup to allow for direct comparison of models initialized with different embeddings.

For each model we extract 6 types of embeddings:
\begin{itemize}
\item Last layer embedding of the first occurrence of a token
\item Average embedding of the last four layers of the first occurrence of a token
\item Last layer embedding of the last occurrence of a token
\item Average embedding of the last four layers of the last occurrence of a token
\item Average of the last layer embedding of every occurrence of a token
\item Average embedding of the last four layers of every occurrence of a token
\end{itemize}

We trained each Siamese network for 100 epochs using the Adam optimizer with a learning rate of $0.001$ to replicate the same setup as \cite{nguyen2021biomedical}. We used a batch size of 8192 in our experiments. The Bi-LSTMs consist of 50 hidden units, the first and second dense layers contain 128 and 50 hidden units respectively. We did run experiments to change the number of hidden units in the dense layers, but found no improvement in performance. We use up to 30 tokens from each atom string and pad the input if needed.

\subsection{UVA with BERT Models}

For each BERT model (i.e., BioBERT, BioBERT Large, BlueBERT, and SapBERT), we predict the synonymy labels for each atom pair in the test set using the input format described in Section \ref{approach}. Results from these experiments are presented in Section \ref{sec:eval}.

\section{Evaluation}
\label{sec:eval}

We evaluate the performance of our models using Accuracy, Precision, Recall, and F-1.


\begin{table}[]
\resizebox{\columnwidth}{!}{
\begin{tabular}{@{}lllrrrrr@{}}
\hline
\textbf{Embedding Type} &  \multicolumn{1}{l}{\textbf{Accuracy}} & \multicolumn{1}{l}{\textbf{Precision}} & \multicolumn{1}{l}{\textbf{Recall}} & \multicolumn{1}{l}{\textbf{F1-Score}} \\ \hline
BioWordVec & 0.9938 & 0.8872 & 0.9274 & 0.9069  \\ \hline
\textbf{SapBERT} & \textbf{0.9886} & \textbf{0.8053} & \textbf{0.854} & \textbf{0.8289}\\
BioBERT & 0.9832 & 0.7232 & 0.7823 & 0.7516 \\
BioBERT\_Large & 0.9854 & 0.7579 & 0.8098 & 0.783 \\
BlueBERT\_Large & 0.9841 & 0.7409 & 0.7833 & 0.7615  \\
UMLSBERT & 0.9834 & 0.7269 & 0.7825	& 0.7537  \\
BioBERT\_Small + SapBERT  & 0.9831 & 0.7202	& 0.7828 &	0.7502 \\
BlueBERT\_Small + SapBERT  & 0.985 & 0.7477 & 0.8121 &	0.7786 \\
UMLSBERT + SapBERT & 0.9842 &	0.7455 &	0.7779 &	0.7614  \\
VanillaBERT + SapBERT & 0.9855 & 0.7637	& 0.7993 & 0.7811  \\ \hline
\end{tabular}
}
\caption{Results for the Siamese Model trained for 100 iterations using BioWordVec embeddings and BERT embeddings extracted using the average token and average of last four layers \cite{nguyen2021biomedical}}
\label{siamese_results_all_models}
\end{table}

\begin{table}[]
\resizebox{\columnwidth}{!}{
\begin{tabular}{@{}lllrrrrr@{}}
\hline
\textbf{Embedding Type} & \multicolumn{1}{l}{\textbf{Accuracy}} & \multicolumn{1}{l}{\textbf{Precision}} & \multicolumn{1}{l}{\textbf{Recall}} & \multicolumn{1}{l}{\textbf{F1-Score}}  \\ \hline
BioWordVec\_Attention & 0.9936 &	0.8884 & 0.9198	& 0.9038  \\ \hline
\textbf{SapBERT\_Attention} & \textbf{0.9886} & \textbf{0.8109} & \textbf{0.8479} & \textbf{0.829} \\
BioBERT\_Attention & 0.9852 & 0.7657 & 0.7823 & 0.7739  \\
BioBERT\_Large\_Attention & 0.9863 & 0.7754 & 0.8128 & 0.7937  \\
BlueBERT\_Large\_Attention & 0.9863 &	0.7754 & 0.8128 & 0.7937 \\
UMLSBERT\_Attention & 0.9851 &	0.7687 &	0.775 &	0.7719  \\
BioBERT\_Small + SapBERT\_Attention & 0.9850 & 0.7634 &	0.7801 & 0.7716 \\
BlueBERT\_Small + SapBERT\_Attention & 0.9863	& 0.7791 &	0.807 & 0.7928  \\
UMLSBERT + SapBERT\_Attention & 0.9848 &	0.7582 &	0.7783 &	0.7681 \\
VanillaBERT + SapBERT\_Attention & 0.9865 & 0.7861 & 0.8014 & 0.7937 \\ \hline
\end{tabular}
}
\caption{Results for the Siamese Model with Attention Layer trained for 100 iterations using BioWordVec embeddings and BERT embeddings extracted using the average token and average of last four layers \cite{nguyenadding}}
\label{siamese_attention_results_all_models}
\end{table}


\begin{table}[]
\resizebox{\columnwidth}{!}{
\begin{tabular}{@{}lrrrrr@{}}
\hline
\textbf{Model\_Type} & \multicolumn{1}{l}{\textbf{Accuracy}} & \multicolumn{1}{l}{\textbf{Precision}} & \multicolumn{1}{l}{\textbf{Recall}} & \multicolumn{1}{l}{\textbf{F1}} \\ \hline
BioBERT & 0.5308 & 0.9683 & 0.5326 & 0.6872  \\
\textbf{BioBERT\_Large} & \textbf{0.6255} & \textbf{0.9870} & \textbf{0.6212} & \textbf{0.7625} \\
BlueBERT & 0.1875 & 0.9985 & 0.1607 & 0.2768 \\
SapBERT & 0.4929 & 0.6528 & 0.4926 & 0.6528  \\
UMLSBERT & 0.5105 &	0.9677 & 0.5113	& 0.6691 \\
BioBERT + SapBERT & 0.5550 & 0.9733 & 0.5554 & 0.7072 \\
BlueBERT+ SapBERT & 0.6261 & 0.9722 & 0.6318 & 0.7658 \\
UMLSBERT + SapBERT & 0.3325 & 0.9714 & 0.3197 & 0.4811  \\
VanillaBERT + SapBERT & 0.5767 & 0.9732 & 0.5785 & 0.7257 \\ \hline
\end{tabular}

}
\caption{Results for Synonymy Prediction using BERT models. Input format: ([CLS] $str_i$ [SEP] $str_j$ [SEP])}
\label{BERT_Synonymy_Prediction}
\end{table}

\begin{table}[]
\resizebox{\columnwidth}{!}{
\begin{tabular}{@{}lrrrrr@{}}
\hline
\textbf{Model\_Type} & \multicolumn{1}{l}{\textbf{Accuracy}} & \multicolumn{1}{l}{\textbf{Precision}} & \multicolumn{1}{l}{\textbf{Recall}} & \multicolumn{1}{l}{\textbf{F1}}  \\ \hline
BioBERT & 0.4202 & 0.9778 & 0.4102 & 0.5780\\
\textbf{BioBERT\_Large} & \textbf{0.6364} & \textbf{0.9864} & \textbf{0.6331} & \textbf{0.7712} \\
BlueBERT & 0.2015 & 0.9984 & 0.1752 & 0.2981  \\
SapBERT & 0.4500 & 0.9668 & 0.4470 & 0.6114 \\
UMLSBERT & 0.4901 & 0.9678 &	0.4900 & 0.6501  \\
BioBERT + SapBERT & 0.5704 & 0.9683 & 0.5750 & 0.7215  \\
BlueBERT+ SapBERT & 0.3576 & 0.9636 & 0.3494 & 0.5128 \\
UMLSBERT + SapBERT & 0.2371 & 0.9598 &	0.2210 & 0.3592 \\
VanillaBERT + SapBERT & 0.4337 &	0.9636 & 0.4311 &	0.5957 \\ \hline
\end{tabular}
}
\caption{Results for Synonymy Prediction using BERT models. Input format: ([CLS] $str_j$ [SEP] $str_i$ [SEP])}
\label{BERT_Synonymy_Prediction_aui2_aui1}
\end{table}

\textbf{Feature Extraction with Siamese Network Results.}  Table \ref{siamese_results_all_models} presents the synonymy prediction using the Siamese Network with embeddings extracted from BERT models and BioWordVec embeddings. Due to space limitations, we only share results for the average token embedding and the average of last four layers of each BERT model. Table \ref{siamese_attention_results_all_models} presents the synonymy prediction using the Siamese Network with an attention layer with embeddings extracted from BERT models and BioWordVec embeddings. Our feature extraction results indicate that averaging all token embeddings and using the average of the last four hidden layers provides the most useful embedding for most models. Additionally, we find that using the Siamese Network with the attention layer achieves better performance in terms of F1-score. Surprisingly, using the embeddings extracted from the biomedical BERT model do not outperform the two baselines of the Siamese Networks with BioWordVec. We find that the embeddings extracted from SapBERT model lead to the best performance for synonymy prediction.

\noindent
\textbf{Synonymy Prediction with BERT Results.} Tables \ref{BERT_Synonymy_Prediction} and \ref{BERT_Synonymy_Prediction_aui2_aui1} present the synonymy prediction results using the BERT architecture. The pre-trained biomedical BERT models do not outperform the current SOTA Siamese Networks for synonymy prediction. We find that the BioBERT Large model is the best performing model for Synonymy prediction. We attribute the performance of this model to its size. We run additional experiments to determine if changing the order of the atoms affects the performance of the models (i.e. feeding ($str_j$, $str_i$) as input instead of ($str_i$, $str_j$)). These results are present in Table \ref{BERT_Synonymy_Prediction_aui2_aui1}. We see that changing the order of the input atom strings only improves the performance for two biomedical BERT models: BioBERT and BlueBert + SapBERT. The results increase for BioBERT and BlueBert + SapBERT by about 0.026\% and 0.049\% in terms of F1-score.

\noindent
\textbf{Discussion.} Our current results for both set of experiments do not outperform the current SOTA. These results indicate that the pre-trained language models on the datasets from the same domain are not enough to accurately predict synonymy among atoms in the Metathesaurus. The number of parameters for the Siamese Networks initialized with BioWordVec are one magnitude higher than all of the BERT models. The large size of the vocabulary could be an indication as to why BioWordVec performs well when compared to the biomedical BERT based variants. Additionally, from our synonymy prediction task we find that using a BERT model trained on the right data and the right task yeilds larger gains in performance for synonymy prediction. The SapBERT model is trained on PubMed and incorporates knowledge from the UMLS Metathesaurus in two ways: 1) using semantic type embeddings and 2) modifying the MLM task to indicate if which words belong to the same concept. These changes to the model likely indicate why it outperforms the other biomedical BERT models for synonymy prediction using the Siamese Networks.
Additionally, we find that using the biomedical BERT embeddings with the Siamese Network yields better results than using the pre-trained BERT models for the UVA task.
These results indicate that further fine-tuning is required to use the biomedical BERT models for the UVA task.

\section{Conclusion}
In this paper, we evaluated different methods of using biomedical BERT-based models in two model architectures: (1) Siamese Network, and (2) Transformer for predicting synonymy  in  the  UMLS  Metathesaurus.  We  aimed  to  validate  if these approaches using the BERT models can actually out-perform the existing approaches. We replace the BioWordVec embeddings in the Siamese Networks with the biomedical BERT embeddings  extracted  from different models  using  different ways of extraction. We evaluate the use of the different pre-trained biomedical BERT models using the transformer architecture that. Our  experiments  yield  surprisingly interesting results: (1) in both model architectures, the approaches employing these biomedical BERT-based models do not outperform  the  existing  approaches  using  Siamese  Network  with BioWordVec  embeddings  for  the  UMLS  synonymy  prediction task, (2) the original BioBERT large model that has not been pre-trained with the UMLS outperforms the SapBERT models that have been pre-trained with the UMLS, and (3) using  the  Siamese  Networks  yields  better  performance  for synonymy prediction when compared to using the biomedical BERT models.


\section{Acknowledgments}
 We would like to thank Nguyen et al. for sharing the training data used in \cite{nguyen2021biomedical, nguyenadding}. We would also like to thank Liu et al. for providing the additional pre-trained SapBERT models (i.e., BioBERT + SapBERT, BlueBERT + SapBERT, UMLSBERT + SapBERT, VanillaBERT + SapBERT) \cite{liu2020self}.


\bibliography{aaai22}
\bibliographystyle{abbrv}

\end{document}